\documentclass[10pt, a4paper]{article}
\usepackage{lrec}
\usepackage{natbib}
\usepackage{multibib}
\newcites{languageresource}{Language Resources}
\usepackage{graphicx}
\usepackage{tabularx}
\usepackage{soul}
\usepackage{titlesec}
\titleformat{\section}{\normalfont\large\bf\center}{\thesection.}{1em}{}
\titleformat{\subsection}{\normalfont\SmallTitleFont\bf\raggedright}{\thesubsection.}{1em}{}
\titleformat{\subsubsection}{\normalfont\normalsize\bf\raggedright}{\thesubsubsection.}{1em}{}
\renewcommand\thesection{\arabic{section}}
\renewcommand\thesubsection{\thesection.\arabic{subsection}}
\renewcommand\thesubsubsection{\thesubsection.\arabic{subsubsection}}

\usepackage{epstopdf}
\usepackage[utf8]{inputenc}

\usepackage{hyperref}
\usepackage{xstring}

\usepackage{color}

\usepackage{tabu}
\usepackage{multirow}
\usepackage{booktabs}
\usepackage{linguex}

\title{Automatic Correction of Syntactic Dependency Annotation Differences}

\name{Andrew Zupon, Andrew Carnie, Michael Hammond, Mihai Surdeanu}

\address{University of Arizona\\
	Tucson, AZ, USA\\
	\{zupon, carnie, hammond, msurdeanu\}@email.arizona.edu\\}

\abstract{
	Annotation inconsistencies between data sets can cause problems for low-resource 
	NLP, where noisy or inconsistent data cannot be as easily 
	replaced compared with resource-rich languages. In this paper, we propose 
	a method for automatically detecting annotation mismatches between dependency parsing 
	corpora, as well as three related methods for automatically converting 
	the mismatches. All three methods rely on comparing an unseen 
	example in a new corpus with similar examples in an 
	existing corpus. These three methods include a simple lexical replacement using 
	the most frequent tag of the example in the existing 
	corpus, a GloVe embedding-based replacement that considers a wider pool 
	of examples, and a BERT embedding-based replacement that uses contextualized 
	embeddings to provide examples fine-tuned to our specific data. We 
	then evaluate these conversions by retraining two dependency parsers---Stanza 
	\citep{qi2020stanza} and Parsing as Tagging (PaT) \citep{vacareanu2020parsing}---on the converted 
	and unconverted data. We find that applying our conversions yields 
	significantly better performance in many cases. Some differences observed 
	between the two parsers are observed. Stanza has a more complex architecture with a quadratic algorithm, so it takes longer to train, but it can generalize better with less data. The PaT parser has a simpler architecture with a linear algorithm, speeding up training time but requiring more training data to reach comparable or better performance.\\ \newline  }

\begin{document}

\newcommand{\todo}[1]{\textcolor{red}{TODO: #1}}
\newcommand{\edit}[1]{\textcolor{blue}{#1}}

\setlength{\tabcolsep}{0.4em} 

\renewcommand{\bibsection}{}

\maketitleabstract

\section{Introduction}

Examples found in one data set that are not found in another can represent either unseen examples or inconsistencies in annotation.
These annotation differences may not be a major problem for resource-rich languages---if a text or data set contains errors or inconsistencies, the researcher can remove the faulty data and replace it with valid data---but this is not true for low-resource languages.
Low-resource languages have limited amounts of annotated data available for NLP tasks, so any data that is compromised by errors or inconsistent annotations cannot as easily be replaced.

\smallskip
This project looks at reducing annotation differences between two different corpora in order to augment training data in a more informed, consistent way.
If successful, this automatic conversion can aid research on low-resource NLP by embiggening the pool of clean, usable data available to researchers.

\smallskip
The contributions of this work are:

\paragraph{(1)} We propose a simple approach for automatically identifying mismatches between two Universal Dependencies (UD) dependency parsing data sets.
First, we identify all the tokens in a head-dependent relation in each data set, along with the specific relations they occur with.
Next, we identify the relations that occur between a head-dependent word pair in the second data set but not the first.
This results in a set of potential annotation errors.

\paragraph{(2)} Once we have our set of potential annotation errors in the second data set, we propose three methods for automatically converting the data.
In the simplest version, we use the most frequent relation for a given head-dependent word pair in the first data set to replace the unseen relation in the second data set.
A more complex approach uses GloVe embeddings \citep{pennington2014glove} to expand the set of head-dependent word pairs in the first data set from which to select the most frequent relation to replace the unseen relation in the second data set.
Our final approach uses BERT embeddings \citep{DBLP:journals/corr/abs-1810-04805} contextualized on our specific training data, but is otherwise identical to our GloVe-based approach.

\smallskip
An example of the lexical version is shown in Table~\ref{table:intro}. 
In this toy example, we see that the word pair $<$such, as$>$ only occurs in Corpus A with the UD relation \texttt{fixed} (for fixed expressions), and it occurs in Corpus B with the UD relations \texttt{mwe} and \texttt{advmod}.
As these two relations are unseen for this word pair in Corpus A, we want to change the entries in Corpus B to more closely match Corpus A.
In this case, all instances of \texttt{mwe} and \texttt{advmod} in Corpus B for the word pair $<$such, as$>$ are replaced with the most frequent label in Corpus A, which is \texttt{fixed}.

\begin{table}[tb!]
\begin{center}
\begin{tabu} to \textwidth {ccc}\toprule
Corpus & Word Pair & Relations \& Counts\\\midrule
A & $<$such, as$>$ & \{fixed: 35\}\\
B (original) & $<$such, as$>$ & \{mwe: 20, advmod: 5\}\\
B (converted) & $<$such, as$>$ & \{fixed: 25\}\\\bottomrule
\end{tabu}
\end{center}
\caption{Lexical replacement approach. We replace unique relations for a word pair in Corpus B with the most frequent relation for that pair in Corpus A.}
\label{table:intro}
\end{table}

\paragraph{(3)} Despite its simplicity, we show that this automatic approach performs well in certain contexts.
For example, the best performing model is a converted condition in all but one case, and many of the converted conditions are significantly better than the unconverted condition.

\section{Related Work}

Our project builds on work in syntactic dependency parsing, data cleaning, and low-resource NLP.


\smallskip
The NLP task used to evaluate our approaches is syntactic dependency parsing.
The core of this task involves identifying the unique syntactic head for a given token and the label of the relation that holds between the head and its modifiers.
The two parsers we use for our evaluation are Stanza \citep{qi2020stanza} and Parsing as Tagging (PaT) \citep{vacareanu2020parsing}.
These parsers are part of an ongoing trend in dependency parsing that marries simplicity with performance.
Details about these two parsers can be found in Section\ \ref{sec:parser}.
Other papers in this direction include \cite{fernandez2019left}, \cite{ma2018stack}, and \cite{kiperwasser2016simple}.


\smallskip
Data cleaning can be a problem within a single source, but becomes especially important when combining data from different sources.
\cite{rahm2000data} and \cite{chu2016data} provide general overviews of data cleaning approaches and challenges.
Specific to NLP, \cite{fu2020rethinking} considers the problem of combining different corpora for a named entity recognition task.
They develop two metrics for measure the similarity between two data sets, then show how that measure correlates with a model's performance on a cross-data-set generalization experiment.
They additionally experiment with detecting and correcting annotation errors in their data sets.
Their approach, however, involves manual correction, as the errors they identify in the named entity recognition data sets are non-systematic and hard to automatically fix.


\smallskip
Lack of annotated training data is one of the hallmarks of a low-resource language.
A resource like word embeddings can be created for a low-resource language based on raw, unannotated text, but syntactic parsing relies on having annotations.
Universal Dependencies (UD), a framework for annotating syntactic and morphological information, has annotated data sets available for over 100 languages, with more being added all the time.
This is inching towards lower-resource languages, but there are still many languages not yet supported.
For this reason, there has been a lot of work on speeding up or even bypassing the annotation process for low-resource languages.
\cite{tiedemann2016bootstrapping} describes an approach for bootstrapping a dependency parser for Maltese (Semitic: Malta) by using annotation projection and model transfer from other languages.
They consider languages close to Maltese by language family or language contact as well as languages with high-performing dependency parsers.
\cite{tiedemann2016tagging} describes an effort to morphologically tag Ingush (Northeast Caucasian: Russia) via interlinear glosses in English from linguistic fieldwork notes.
The results of these approaches is promising, but the authors note that it may be more practical at times to invest in manual annotation than to try to tweak transfer models.

\smallskip
Our project builds upon these three categories of work.
For dependency parsing, we compare two recent parsers part of the ongoing trend towards simplicity.
For data cleaning, we propose automatic correction methods that streamline the process compared to previous manual corrections.
For low-resource languages, we demonstrate that our automatic methods can improve performance on parsing without the need to manually annotate additional data.

\section{Approach}

This section discusses our method for automatically identifying syntactic dependency annotation differences between two corpora and our three approaches for automatically creating the converted training data sets used for dependency parsing using those annotation differences.

\subsection{Identifying Annotation Differences}

All three approaches to converting the augment corpus data first require that we identify annotation differences between the data sets that may need converting.
The method of identifying these annotation differences follows.
First, we collect a list of all the head-dependent word pairs in the base corpus (the corpus we will use for testing), along with all the relations that occur with each of those pairs.
Then, we collect similar lists of all the head-dependent word pairs in the augment corpus (the corpus we will add to the base corpus) and the relations that occur with those pairs.
For the purposes of this project, any relation that occurs with a head-dependent word pair in the augment corpus but \textit{not} in the base corpus is an annotation difference.

\smallskip
Once these differences are are identified, we proceed to the next step of automatically conversions.
For this project, the three approaches for automatically converting the data use the same set of annotation differences identified with the method described in this section.

\subsection{Lexical Approach}


The first approach uses a na\"ive token-based method of replacement.
For a head-dependent-relation triple from the augment corpus that doesn't show up in the base corpus, we simply replace the relation for that specific head-dependent-relation triple in the augment corpus with the most common relation for that head-dependent word pair in the base corpus.
This is essentially retagging with the mot common tag, but only in cases where the relation for the word pair is unobserved in the augment corpus.
From our toy example in Table~\ref{table:intro}, this would mean replacing \texttt{mwe} and \texttt{advmod} with \texttt{fixed} for the word pair $<$such, as$>$, as \texttt{fixed} is the most common tag for that pair in the base corpus.
The benefit of this approach is that it is simple to implement and it does not require word embeddings for the language.

\subsection{GloVe Embedding Approach}

\begin{table}[tb!]
\begin{center}
\begin{tabu} to \textwidth {ccc}\toprule
Corpus & Word Pair & Relations \& Counts\\\midrule
A & $<$have, n't$>$ & \{dep: 9\}\\
A & $<$has, n't$>$ & \{neg: 5\}\\
A & $<$would, n't$>$ & \{neg: 5\}\\
B (original) & $<$have, n't$>$ & \{advmod: 6\}\\
B (converted) & $<$have, n't$>$ & \{neg: 6\}\\\bottomrule
\end{tabu}
\end{center}
\caption{Embedding-based replacement approach. The most frequent relation when considering only the exact pair $<$have, n't$>$ is the (incorrect) \texttt{dep}. The most frequent relation when considering the exact pair \textit{and} related word pairs using vector similarity is the (correct) \texttt{neg}.} 
\label{table:glove}
\end{table}

As with the Lexical approach, we want to replace the unseen relation in the augment corpus with a relation we have seen in the base corpus.
However, instead of relying on the exact head-dependent word pair, which can be sparse, this approach uses GloVe embeddings \citep{pennington2014glove} to generalize to additional word pairs.
For each word in the head-dependent word pair, we use the Pymagnitude \citep{patel-etal-2018-magnitude} library for Python with GloVe vectors to generate the top 10\footnote{This is a hyperparameter that could be trained or adjusted. Also note that the similarity scores for the top 10 most similar words were not filtered below any threshold.} most similar words to the original word.
From these candidates, we create new candidate word pairs by combining each new head and dependent word.
From this set of candidate word pairs, we then see which ones actually occur in the base corpus.
Out of these, plus the original word-pair, we then choose the most frequent relation overall to replace the relation for the original head-dependent word pair in the augment corpus.
An example of this is shown in Table~\ref{table:glove}.
In this case, selecting from a wider set of word pairs yields a better replacement (\texttt{neg}) than only considering the exact match (\texttt{dep}).
This approach has the benefit of being able to generalize beyond the exact word pair, which can be especially useful in low-resource settings with limited data.
However, it does rely on having pretrained word embeddings available for the language.

\subsection{BERT Embedding Approach}

This approach is similar to the GloVe Embedding approach, but it uses a different strategy for generating word embeddings.
Unlike the GloVe vectors, which are pretrained on data that could differ from the specific data one is working with, BERT embeddings are contextualized based on the specific texts you provide.
For this approach, we generate contextualized BERT embeddings using a multilingual pretrained BERT model \citep{DBLP:journals/corr/abs-1810-04805} for each sample for each training data partition of the base corpus.
We then use these new BERT embeddings to generate the new candidate word pairs.
Table~\ref{table:glove} applies to this approach as well, but the pool of candidate word pairs may differ due to the different embeddings used to generate them.
Like with the GloVe approach, this has the benefit of generalizing beyond the exact word pair.
One drawback of this approach is the need for a pretrained BERT model.
There are multilingual BERT models available, but the largest only covers 104 languages.

\section{Experimental Setup}

This section describes the experimental settings we use for evaluating our automatic conversions.
This includes a discussion of the data sets used, the amounts of training data used to simulate different low-resource conditions, the two parsers used for training, and information about our training conditions.

\subsection{Data Sets}

Limited data can have a negative effect on performance, as the model may not have seen enough examples to generalize well.
Being able to leverage additional data to help train a new dependency parser could help improve parsing performance.
However, when that data is inconsistent with the original training data, problems can be increased instead of alleviated.
One limiting factor is that to do a comparison between data sets, we need a language to have more than one data set available.
This unfortunately excludes the lowest-resource of low-resource languages.

\smallskip
The data sets chosen for this experiment follow the Universal Dependencies\footnote{\url{https://universaldependencies.org/}} framework.
Every data set should be marked up using a consistent annotation scheme, but some variation exists.
For example, there are different versions of the Universal Dependencies annotations, and some data sets are manually created while others are automatically converted from other treebanks.

\smallskip
For this test case, we only consider English as a proof of concept, but this approach could easily be extended to actual low-resource languages.
For the base corpus, we use the Georgetown University Multilayer (GUM) corpus \citep{zeldes2017gum}.
For the augment corpus, we use the Wall Street Journal (WSJ) portion of the Penn Treebank \citep{taylor2003penn} converted into the \texttt{conllu} format used for Universal Dependencies data.
Information about these two corpora is shown in Table~\ref{table:depcorpora}.

\begin{table}[tb]
	\begin{center}
		
		\begin{tabu} to \textwidth {ccccc}\toprule
			{Corpus} & {Total} & {Train} & {Dev} & {Test}\\\midrule
			GUM & 5961 & 4287 & 784 & 890\\
			WSJ & 47 287 & 39 832 & 5039 & 2416\\\bottomrule
		\end{tabu}
	\end{center}
	\caption{Number of sentences in each partition of the GUM and WSJ corpora.}
	\label{table:depcorpora}
\end{table}

\subsection{Training Data Amount}

The amount of data used when training a parser can affect its performance.
More data often leads to better performance, but the rate of improvement can vary depending on the type of parser.
We consider a spectrum of training data amounts to simulate different low-resource settings.
For this experiment, we use training amount of 250, 500, 1000, 2000, and 4000 sentences.

\smallskip
For each amount, half of the sentences come from the base corpus (GUM) and half come from the augment corpus (WSJ).
To generalize better, for each training data amount we sample three times from each corpus.
For example, for the 1000 sentence training amount we sample 500 sentences from the GUM training partition and 500 sentences from the WSJ training partition.
We repeat this process two more times in order to have three runs to compare for each training data amount.

\subsection{Choice of Parser}
\label{sec:parser}

The specific architecture of the parser used can interact with the amount of training data to affect performance.
Some parsers need more training data to generalize well, whereas others can generalize from less data.
In order to explore how different parsers can affect performance, we consider two neural-based dependency parsers:\ \ Stanford's Stanza parser \cite{qi2020stanza}, and \citet{vacareanu2020parsing}'s Parsing as Tagging (PaT) parser.

\smallskip
Neural networks require large amounts of training data, so neural network-based dependency parsers perform better with large amounts of data.
These same neural parsers are also known to show a drop in performance when they don't have a large training data set relative to rule-based parsers (Kabiri, p.c.).
While this project only considers neural-based parsers, future work could compare how a rule-based parser performs in similar low-resource training conditions.

\smallskip
It is also important to note that this project is not about modifying or improving the dependency parsers themselves.
Rather, we use existing parsers as-is to investigate how data augmentation and conversion methods can help improve dependency parsing performance.
Hence, we do not expect to reach or exceed the performance of any newer state-of-the-art models.

\subsubsection{Stanza Parser}

Stanza \citep{qi2020stanza} is a multilingual open-source Python NLP toolkit.
It features a fully neural text analysis pipeline that supports tokenization, lemmatization, part-of-speech and morphological tagging, dependency parsing, and named entity recognition.
The Stanza dependency parser is the Bi-LSTM-based deep biaffine neural dependency parser developed by \cite{dozat2016deep} augmented with additional linguistically motivated features.
One new feature predicts the linearization order of two words in a given language, and the other new feature predicts the typical distance in linear order between them.
This results in a quadratic algorithm, since it combines the embedding of the modifier with the embedding for each possible head in the sentence.
As we will see, this results in overall higher and more consistent performance, even with less training data, but does increase complexity and runtime.

\smallskip
\cite{dozat2016deep}'s graph-based dependency parser extends earlier work from \cite{kiperwasser2016simple}.
These extensions include a larger network with more regularization, using a biaffine attention mechanism and label classifier instead of an affine one, and reducing the dimensionality of the top recurrent states of the LSTM by putting them through MLP operations before using them in the biaffine transformations.
These modifications keep the simplicity of neural approaches while approaching transition-based parser performance.

\subsubsection{Parsing as Tagging (PaT) Parser}

The Parsing as Tagging (PaT) parser \citep{vacareanu2020parsing} treats dependency parsing as a sequence model using a bidirectional LSTM over BERT embeddings.
In this case, the tag that is predicted for each token is the relative position of that token's syntactic head.
This reframing of dependency parsing into a sequence tagging task that relies only on surface information, rather than syntactic structure, simplifies dependency parsing without compromising on performance (at least when plenty of training data is available).
The PaT parser reaches state-of-the-art or comparable on 12 Universal Dependencies languages compared to previous state-of-the-art performance by \cite{fernandez2019left}.
Unlike Stanza, the PaT algorithm is linear, as it predicts the relative position of the head only based on the embedding of the modifier.
As we will see, while this makes the parser more efficient and quicker to train, it does need more training data in order to generalize, because it relies on less information to predict the head.

\subsection{Training Setup}

With each parser, we trained each of four conditions (Unconverted\footnote{The half of the training data from the augment corpus gets added to the base corpus without any conversions.}, Converted-Lexical, Converted-GloVe, and Converted-BERT) on each of five training data amounts (250, 500, 1000, 2000, and 4000 sentences).
For each training data amount, we trained three times using three different samples from the original training data.
We then tested each of these models on the same original GUM corpus test partition.

\section{Results}
\label{sec:results}

This section discusses the results of parsing with both the Stanza and PaT parsers and includes a short error analysis.

\subsection{Evaluation Metrics}

In this experiment, we report results for two evaluation metrics:\ \ Unlabeled Attachment Score (UAS) and Labeled Attachment Score (LAS).
Unlabeled Attachment Score is based solely on correctly identifying the head word and dependent word without considering the label.
Labeled Attachment Score is based on identifying the head word and dependent word with the correct label.
Based on my method of identifying and converting potential problems, which does not change anything about heads or dependents, we might expect the Unlabeled Attachment Score not to change.
However, due to the architecture of the parsers, where the prediction of the head position and its label are modeled jointly, we do see changes in UAS.
Statistical significance for each condition is calculated based on the median performing model of the three samples based on LAS.

\subsection{Stanza Results}

Table\ \ref{table:results-stanza} shows unlabeled and labeled accuracies using the Stanza parser.

\begin{table*}[tb!]
	\footnotesize
	\begin{center}
		\begin{tabu} to \textwidth {cllllllll} \toprule
			\multirow{2}{*}{Sentences} & \multicolumn{2}{c}{Unconverted} & \multicolumn{2}{c}{Converted-Lexical} & \multicolumn{2}{c}{Converted-GloVe} &\multicolumn{2}{c}{Converted-BERT} \\\cmidrule(lr){2-3}\cmidrule(lr){4-5}\cmidrule(lr){6-7}\cmidrule(lr){8-9}
			
			{}	&	\multicolumn{1}{c}{UAS}	&	\multicolumn{1}{c}{LAS}	&	\multicolumn{1}{c}{UAS}	&	\multicolumn{1}{c}{LAS}	&	\multicolumn{1}{c}{UAS}	&	\multicolumn{1}{c}{LAS}	&	\multicolumn{1}{c}{UAS}	&	\multicolumn{1}{c}{LAS}\\\midrule
			250		&	73.18 \scriptsize{\textpm\ 1.69}	&	67.69 \scriptsize{\textpm\ 2.78}	&	\textbf{75.56 \scriptsize{\textpm\ 0.79}} *	&	\textbf{70.71 \scriptsize{\textpm\ 0.88}} *	&	72.93 \scriptsize{\textpm\ 1.79} *	&	66.13 \scriptsize{\textpm\ 3.57} *	&	74.09 \scriptsize{\textpm\ 2.66} *	&	68.08 \scriptsize{\textpm\ 4.60} *\\
			
			500	&	74.67 \scriptsize{\textpm\ 2.11}	&	69.50 \scriptsize{\textpm\ 2.85}	&	76.97 \scriptsize{\textpm\ 5.63} *	&	72.48 \scriptsize{\textpm\ 6.55} *	&	\textbf{79.46 \scriptsize{\textpm\ 0.71}} *	&	\textbf{75.36 \scriptsize{\textpm\ 0.67}} *	&	\textbf{79.46 \scriptsize{\textpm\ 0.71}} *	&	\textbf{75.36 \scriptsize{\textpm\ 0.67}} *\\
			
			1000	&	73.88 \scriptsize{\textpm\ 3.97}	&	67.77 \scriptsize{\textpm\ 6.14}	&	74.06 \scriptsize{\textpm\ 1.46} *	&	63.92 \scriptsize{\textpm\ 1.21} *	&	\textbf{76.64 \scriptsize{\textpm\ 3.62}} *	&	\textbf{72.21 \scriptsize{\textpm\ 5.11}} *	&	74.28 \scriptsize{\textpm\ 1.53} *	&	69.12 \scriptsize{\textpm\ 2.43} *\\
			
			2000	&	72.03 \scriptsize{\textpm\ 0.86}	&	66.46 \scriptsize{\textpm\ 0.77}	&	72.64 \scriptsize{\textpm\ 1.07} *	&	64.42 \scriptsize{\textpm\ 0.67}	&	76.01 \scriptsize{\textpm\ 4.76} *	&	71.91 \scriptsize{\textpm\ 5.49} *	&	\textbf{79.74 \scriptsize{\textpm\ 4.94}} *	&	\textbf{76.10 \scriptsize{\textpm\ 5.97}} *\\
			
			4000	&	75.74 \scriptsize{\textpm\ 5.69}	&	71.52 \scriptsize{\textpm\ 6.99}	&	74.75 \scriptsize{\textpm\ 0.64}	&	67.29 \scriptsize{\textpm\ 0.53}	&	\textbf{76.57 \scriptsize{\textpm\ 5.16}}	&	\textbf{72.41 \scriptsize{\textpm\ 6.21}} *	&	74.13 \scriptsize{\textpm\ 1.07}	&	68.59 \scriptsize{\textpm\ 1.05}\\\bottomrule
			
		\end{tabu}
		\caption{Accuracy using Stanza. The best condition for each training amount is indicated in \textbf{bold}. * indicates statistical significance at $p<0.05$ between the Converted condition and the Unconverted condition.}
		\label{table:results-stanza}
	\end{center}
\end{table*}

\smallskip
For UAS, a Converted condition outperforms the Unconverted condition at all training data amounts.
Converted-Lexical is the best performing condition for 250 sentences.
Converted-GloVe is the best performing condition or tied for the best for 500, 1000, and 4000 sentences.
Converted-BERT is the best performing condition or tied for the best for 500 and 2000 sentences.
This performance is statistically significant for all training data amounts except for 4000 sentences.

\smallskip
For LAS, a Converted condition outperforms the Unconverted condition at all training data amounts.
Like with UAS, Converted-Lexical is the best performing condition for 250 sentences, Converted-GloVe is best or tied for best for 500, 1000, and 4000 sentences, and Converted-BERT is best or tied for the best for 500 and 2000 sentences.
Unlike with UAS, Converted-Lexical is only statistically significant at 250, 500, and 1000 sentences; Converted-GloVe is statistically significant at all training data amounts; and Converted-BERT is statistically significant at 250, 500, 1000, and 2000 sentences.

\subsection{PaT Results}

Table\ \ref{table:results-pat} shows the unlabeled and labeled accuracies using the PaT parser.

\begin{table*}[tbh!]
	\footnotesize
	\begin{center}
		\begin{tabu} to \textwidth {cllllllll} \toprule
			\multirow{2}{*}{Sentences} & \multicolumn{2}{c}{Unconverted} & \multicolumn{2}{c}{Converted-Lexical} & \multicolumn{2}{c}{Converted-GloVe} &\multicolumn{2}{c}{Converted-BERT} \\\cmidrule(lr){2-3}\cmidrule(lr){4-5}\cmidrule(lr){6-7}\cmidrule(lr){8-9}
			
			{}	&	\multicolumn{1}{c}{UAS}	&	\multicolumn{1}{c}{LAS}	&	\multicolumn{1}{c}{UAS}	&	\multicolumn{1}{c}{LAS}	&	\multicolumn{1}{c}{UAS}	&	\multicolumn{1}{c}{LAS}	&	\multicolumn{1}{c}{UAS}	&	\multicolumn{1}{c}{LAS}\\\midrule
			250		&	40.63 \scriptsize{\textpm\ 13.91}	&	17.08 \scriptsize{\textpm\ 22.51}	&	40.16 \scriptsize{\textpm\ 13.09}	&	16.57 \scriptsize{\textpm\ 21.63}	&	39.91 \scriptsize{\textpm\ 12.67}	&	16.74 \scriptsize{\textpm\ 21.92}	&	\textbf{40.67 \scriptsize{\textpm\ 13.98}}	&	\textbf{17.36 \scriptsize{\textpm\ 23.00}}\\
			
			500	&	64.32 \scriptsize{\textpm\ 3.08}	&	50.23 \scriptsize{\textpm\ 3.92}	&	\textbf{64.36 \scriptsize{\textpm\ 3.81}}	&	50.67 \scriptsize{\textpm\ 4.66} *	&	64.33 \scriptsize{\textpm\ 3.60}	&	\textbf{50.76 \scriptsize{\textpm\ 4.12}}	&	64.33 \scriptsize{\textpm\ 3.60}	&	\textbf{50.76 \scriptsize{\textpm\ 4.12}}\\
			
			1000	&	\textbf{72.33 \scriptsize{\textpm\ 1.50}}	&	61.52 \scriptsize{\textpm\ 2.90}	&	72.30 \scriptsize{\textpm\ 1.66}	&	\textbf{61.58 \scriptsize{\textpm\ 3.61}} *	&	72.09 \scriptsize{\textpm\ 1.38}	&	61.24 \scriptsize{\textpm\ 2.67}	&	71.86 \scriptsize{\textpm\ 1.73}	&	60.99 \scriptsize{\textpm\ 3.12}\\
			
			2000	&	74.64 \scriptsize{\textpm\ 0.63}	&	66.29 \scriptsize{\textpm\ 0.46}	&	74.19 \scriptsize{\textpm\ 0.93}	&	65.18 \scriptsize{\textpm\ 2.30} *	&	74.73 \scriptsize{\textpm\ 1.76} *	&	66.40 \scriptsize{\textpm\ 3.32} *	&	\textbf{74.77 \scriptsize{\textpm\ 1.20}}	&	\textbf{66.97 \scriptsize{\textpm\ 1.07}} *\\
			
			4000	&	81.13 \scriptsize{\textpm\ 0.51}	&	74.44 \scriptsize{\textpm\ 1.19}	&	81.25 \scriptsize{\textpm\ 0.39} *	&	74.20 \scriptsize{\textpm\ 1.54}	&	\textbf{81.64 \scriptsize{\textpm\ 0.64}} *	&	\textbf{74.73 \scriptsize{\textpm\ 1.76}}	&	81.19 \scriptsize{\textpm\ 0.84}	&	74.26 \scriptsize{\textpm\ 1.65}\\\bottomrule
			
		\end{tabu}
		\caption{Accuracy using PaT. The best condition for each training amount is indicated in \textbf{bold}. * indicates statistical significance at $p<0.05$ between the Converted condition and the Unconverted condition.}
		\label{table:results-pat}
	\end{center}
\end{table*}

For UAS, the Unconverted condition outperforms all Converted conditions for with 1000 sentences---the only condition across parsers, training data amounts, and evaluation metrics where the Unconverted condition performs best.
Converted-Lexical is the best performing condition for 500 sentences.
Converted-GloVe is the best performing condition for 4000 sentences.
Converted-BERT is the best performing condition for 250 and 2000 sentences.
Unlike with Stanza, few of these conditions are statistically significant.
Only Converted-Lexical for 4000 sentences and Converted-GloVe for 2000 and 4000 sentences perform significantly better than the Unconverted condition.

\smallskip
For LAS, Converted-Lexical is the best performing condition for 1000 sentences.
Converted-GloVe is the best performing condition or tied for the best for 500 and 4000 sentences.
Converted-BERT is the best performing condition or tied for the best for 250, 500, and 2000 sentences.
More conditions show statistical significance with LAS compared to UAS.
For LAS, Converted-Lexical performs significantly better than Unconverted for 500, 1000, and 2000 sentences, and Converted-GloVe and Converted-BERT perform significantly better than Unconverted for 2000 sentences.

\subsection{Prediction Analysis}

We perform a simple prediction analysis on the best models for each parser.
This involves comparing the predictions of the Unconverted and Converted conditions with the labels in the gold testing data.
The examples reported here only come from sentences where either the Unconverted or Converted predictions differ from the gold data, but not where both Unconverted or Converted predictions are incorrect.
This helps us identify where our Converted model uniquely over- or under-performs relative to the Unconverted model.

\smallskip
For our best Stanza model (Converted-BERT-2000), Table~\ref{table:error-stanza} shows the relations from gold that were most often incorrectly predicted in the Unconverted and Converted conditions\footnote{An incorrect prediction for e.g.\ \texttt{nmod} means that the correct label in gold is \texttt{nmod}, but our model predicted a different label.}.
This table only includes those relations that were predicted incorrectly more than 50 times.
Table~\ref{table:error-pat} shows the same for our best PaT model (Converted-GloVe-4000).
We discuss these prediction errors in more detail in Section~\ref{sec:discussion}.

\begin{table}[tb!]
\begin{center}
\begin{tabu} to \textwidth {cccc}\toprule
Gold & Incorrect & Most Frequent	&	\multirow{2}{*}{Count}\\
Relation	&	Predictions	&	Incorrect Label	&	{}\\\midrule
\multicolumn{4}{c}{Unconverted}\\\cmidrule(lr){1-4}
acl	&	67	&	advcl	&	26	\\
acl:relcl	&	50	&	advcl	&	29	\\
advcl	&	95	&	xcomp	&	26	\\
amod	&	85	&	compound	&	40	\\
appos	&	73	&	nmod	&	29	\\
compound	&	127	&	nmod	&	54	\\
conj	&	184	&	nmod	&	37	\\
flat	&	99	&	compound	&	35	\\
nmod	&	96	&	obl	&	51	\\
nsubj	&	59	&	obj	&	18	\\
nsubj:pass	&	68	&	nsubj	&	55	\\
obj	&	62	&	dobj	&	32	\\
obl	&	207	&	nmod	&	173	\\
root	&	89	&	advcl	&	28	\\\cmidrule(lr){1-4}
\multicolumn{4}{c}{Converted}\\\cmidrule(lr){1-4}
conj	&	51	&	appos	&	32	\\
nmod	&	84	&	obl	&	55	\\
nsubj	&	84	&	nsubj:pass	&	30	\\
obj	&	61	&	obl	&	32	\\
obl	&	65	&	nmod	&	53	\\\bottomrule
\end{tabu}
\caption{Incorrect predictions with our best Stanza model (Converted-BERT-2000) compared to the Unconverted model with the same training amount and random seed. Gold Relation is the label in the gold testing data that is incorrectly predicted with our models over 50 times. Incorrect Predictions is the total number of times our model incorrectly predicted that relation. Most Frequent Incorrect Label is the most common label chosen for the incorrectly predicted gold label. Count is the number of times our model predicts that most frequent incorect label.}
\label{table:error-stanza}
\end{center}
\end{table}

\begin{table}[tb!]
\begin{center}
\begin{tabu} to \textwidth {cccc}\toprule
Gold & Incorrect & Most Frequent	&	\multirow{2}{*}{Count}\\
Relation	&	Predictions	&	Incorrect Label	&	{}\\\midrule
\multicolumn{4}{c}{Unconverted}\\\cmidrule(lr){1-4}
acl	&	56	&	advcl	&	19	\\
advcl	&	75	&	acl	&	17	\\
amod	&	114	&	compound	&	48	\\
appos	&	72	&	conj	&	27	\\
compound	&	255	&	amod	&	74	\\
conj	&	79	&	compound	&	16	\\
flat	&	176	&	conj	&	58	\\
nmod	&	241	&	obl	&	166	\\
nsubj	&	107	&	obj	&	31	\\
obl	&	160	&	nmod	&	77	\\
root	&	77	&	conj	&	33	\\\cmidrule(lr){1-4}
\multicolumn{4}{c}{Converted}\\\cmidrule(lr){1-4}
compound	&	176	&	amod	&	117	\\
nmod	&	145	&	obl	&	103	\\
nsubj	&	92	&	conj	&	27	\\
obj	&	117	&	conj	&	24	\\
root	&	60	&	advcl	&	16	\\\bottomrule
\end{tabu}
\caption{Incorrect predictions with our best PaT model (Converted-GloVe-4000) compared to the Unconverted model with the same training amount and random seed. Gold Relation is the label in the gold testing data that is incorrectly predicted with our models over 50 times. Incorrect Predictions is the total number of times our model incorrectly predicted that relation. Most Frequent Incorrect Label is the most common label chosen for the incorrectly predicted gold label. Count is the number of times our model predicts that most frequent incorect label.}
\label{table:error-pat}
\end{center}
\end{table}


\subsection{Discussion}
\label{sec:discussion}

The results above show a few trends.
First, we find that our conversions work for both parsers.
Going through the process of converting the data is worthwhile in many cases.
For Stanza, performing any conversion results in significantly higher performance in 12/15 cases for UAS and 12/15 cases for LAS.
For PaT, converting the data yields significantly higher performance in more limited cases---only 3/15 for UAS and 5/15 for LAS.

\smallskip
Another finding is that in general, matching things semantically---that is, using word vectors---is better than a simple lexical match.
With Stanza, there is only one training amount (250 sentences) where Converted-GloVe or Converted-BERT is not the best performing model.
Similarly, there are only two training amounts for PaT (500 and 1000 sentences) where Converted-GloVe or Converted-BERT is not the best performing model.
In addition, we observe that using BERT specifically can yield the best improvements.
Our overall best Stanza model uses BERT embeddings, and our best performing condition across parsers is a Converted-BERT model for 4/10 training amounts for UAS\footnote{2/5 for Stanza, 2/5 for PaT} and 5/10 training amounts for LAS\footnote{2/5 for Stanza, 3/5 for PaT}.

\smallskip
We also see a striking difference between the Stanza and PaT parsers.
Sanza performs consistently across different training data amounts.
The difference in performance for between the best and worst performing Unconverted condition with Stanza is 2.56 for UAS and 5.06 for LAS.
Contrast this with the PaT parser, where the difference between the best and worst performing Unconverted condition with PaT is 40.5 for UAS and 57.36 for LAS.
This is likely due to the architectures of each parser.
As discussed previously, the Stanza algorithm is more complex, and can generalize from less data than PaT.
Thus, we see PaT lagging in performance in training conditions with less data.
However, we also see that in higher-data conditions (2000 and 4000 sentences), PaT begins to outperform Stanza.

\smallskip
Finally, our prediction analysis reveals areas where our Converted models improve over the Unconverted models and some remaining areas for improvement.
Compared with the Unconverted models, we observe a reduction in the number of relation types predicted incorrectly with the Converted models.
Stanza Unconverted predicts 14 relations incorrectly more than 50 times, and PaT Unconverted predicts 11 relations incorrectly more than 50 times.
Both Converted models reduce this down to only five relations incorrectly predicted more than 50 times.
We also see a reduction in the overall count of incorrect predictions for each relation type, and we notice no cases where the Converted model incorrectly predicts a relation more than 50 times where the Unconverted model does not.
That is, the Converted models do not seem to be introducing new classes of errors.
However, there are still areas to improve.
There are some labels (notably \texttt{nmod}, \texttt{nsubj}, and \texttt{obj}) that our best models with both parsers incorrectly predict more than 50 times.
One explanation for these errors is that the relations involved are syntactically and semantically similar.
For example, \texttt{nsubj} and \texttt{nsubj:pass} both involve clausal subjects, \texttt{nmod} and \texttt{obl} both involve nominal phrase modifiers, and \texttt{compound} and \texttt{amod} both involve multiword expressions.

\smallskip
The results of our parsing experiment and prediction analysis suggest that applying our simple, automatic conversion methods to the training data can result in a model that outperforms a simpler model that does not utilize our proposed methods with very little additional time or labor needed.

\section{Conclusion}

We proposed a method for automatically identifying mismatches between two Universal Dependencies dependency parsing corpora and proposed three related approaches for automatically converting the data.
We then retrained two different dependency parsers with the converted data to evaluate how these methods perform compared to an unconverted baseline with different amounts of training data to simulate low-resource conditions.
Despite differences between the two parsers, we find that our approaches yield significantly better performance in many conditions compared to the baseline.
This work suggests that automatically identifying and converting mismatches between two data sets can serve as a simple way to augment limited training data and improve dependency parsing performance in low-resource scenarios.

\smallskip
For reproducibility, we release the code behind this work as open source.
The software is available at this URL:\\
\url{https://github.com/clulab/releases/tree/master/lrec2022-parsing}.

\section{Acknowledgments}\label{sec:ack}

We gratefully thank Roya Kabiri and Maria Alexeeva for their insightful feedback and help setting up the two parsers and other software.
This work was supported by the Defense Advanced Research Projects Agency (DARPA) under the World Modelers program, grant number W911NF1810014.

\section{Bibliographical References}\label{sec:references}

\bibliographystyle{apalike}
\bibliography{lrec-parsing}

\begin{thebibliography}{}

\bibitem[Chu et~al., 2016]{chu2016data}
Chu, X., Ilyas, I.~F., Krishnan, S., and Wang, J. (2016).
\newblock Data cleaning: Overview and emerging challenges.
\newblock In {\em Proceedings of the 2016 international conference on
  management of data}, pages 2201--2206.

\bibitem[Devlin et~al., 2018]{DBLP:journals/corr/abs-1810-04805}
Devlin, J., Chang, M., Lee, K., and Toutanova, K. (2018).
\newblock {BERT:} pre-training of deep bidirectional transformers for language
  understanding.
\newblock {\em CoRR}, abs/1810.04805.

\bibitem[Dozat and Manning, 2016]{dozat2016deep}
Dozat, T. and Manning, C.~D. (2016).
\newblock Deep biaffine attention for neural dependency parsing.
\newblock {\em arXiv preprint arXiv:1611.01734}.

\bibitem[Fern{\'a}ndez-Gonz{\'a}lez and G{\'o}mez-Rodr{\'\i}guez,
  2019]{fernandez2019left}
Fern{\'a}ndez-Gonz{\'a}lez, D. and G{\'o}mez-Rodr{\'\i}guez, C. (2019).
\newblock Left-to-right dependency parsing with pointer networks.
\newblock {\em arXiv preprint arXiv:1903.08445}.

\bibitem[Fu et~al., 2020]{fu2020rethinking}
Fu, J., Liu, P., Zhang, Q., and Huang, X. (2020).
\newblock Rethinking generalization of neural models: A named entity
  recognition case study.
\newblock In {\em AAAI}, pages 7732--7739.

\bibitem[Kiperwasser and Goldberg, 2016]{kiperwasser2016simple}
Kiperwasser, E. and Goldberg, Y. (2016).
\newblock Simple and accurate dependency parsing using bidirectional lstm
  feature representations.
\newblock {\em Transactions of the Association for Computational Linguistics},
  4:313--327.

\bibitem[Ma et~al., 2018]{ma2018stack}
Ma, X., Hu, Z., Liu, J., Peng, N., Neubig, G., and Hovy, E. (2018).
\newblock Stack-pointer networks for dependency parsing.
\newblock {\em arXiv preprint arXiv:1805.01087}.

\bibitem[Patel et~al., 2018]{patel-etal-2018-magnitude}
Patel, A., Sands, A., Callison-Burch, C., and Apidianaki, M. (2018).
\newblock {M}agnitude: A fast, efficient universal vector embedding utility
  package.
\newblock In {\em Proceedings of the 2018 Conference on Empirical Methods in
  Natural Language Processing: System Demonstrations}, pages 120--126,
  Brussels, Belgium. Association for Computational Linguistics.

\bibitem[Pennington et~al., 2014]{pennington2014glove}
Pennington, J., Socher, R., and Manning, C.~D. (2014).
\newblock Glove: Global vectors for word representation.
\newblock In {\em Proceedings of the 2014 conference on empirical methods in
  natural language processing (EMNLP)}, pages 1532--1543.

\bibitem[Qi et~al., 2020]{qi2020stanza}
Qi, P., Zhang, Y., Zhang, Y., Bolton, J., and Manning, C.~D. (2020).
\newblock Stanza: A {Python} natural language processing toolkit for many human
  languages.
\newblock In {\em Proceedings of the 58th Annual Meeting of the Association for
  Computational Linguistics: System Demonstrations}.

\bibitem[Rahm and Do, 2000]{rahm2000data}
Rahm, E. and Do, H.~H. (2000).
\newblock Data cleaning: Problems and current approaches.
\newblock {\em IEEE Data Eng. Bull.}, 23(4):3--13.

\bibitem[Taylor et~al., 2003]{taylor2003penn}
Taylor, A., Marcus, M., and Santorini, B. (2003).
\newblock The penn treebank: an overview.
\newblock {\em Treebanks}, pages 5--22.

\bibitem[Tiedemann et~al., 2016]{tiedemann2016tagging}
Tiedemann, J., Nichols, J., and Sprouse, R. (2016).
\newblock Tagging ingush-language technology for low-resource languages using
  resources from linguistic field work.
\newblock In {\em Proceedings of the Workshop on Language Technology Resources
  and Tools for Digital Humanities (LT4DH)}, pages 148--155.

\bibitem[Tiedemann and van~der Plas, 2016]{tiedemann2016bootstrapping}
Tiedemann, J. and van~der Plas, L. (2016).
\newblock Bootstrapping a dependency parser for maltese--a real-world test
  case.
\newblock {\em From Semantics to Dialectometry: Festschrift in honor of John
  Nerbonne. College Publications, London}.

\bibitem[Vacareanu et~al., 2020]{vacareanu2020parsing}
Vacareanu, R., Barbosa, G. C.~G., Valenzuela-Escarcega, M.~A., and Surdeanu, M.
  (2020).
\newblock Parsing as tagging.
\newblock In {\em Proceedings of The 12th Language Resources and Evaluation
  Conference}, pages 5225--5231.

\bibitem[Zeldes, 2017]{zeldes2017gum}
Zeldes, A. (2017).
\newblock The gum corpus: Creating multilayer resources in the classroom.
\newblock {\em Language Resources and Evaluation}, 51(3):581--612.

\end{thebibliography}

\end{document}